\documentclass{article}
\usepackage{ijcai21}

\usepackage{times}
\usepackage{url}
\usepackage[hidelinks]{hyperref}
\usepackage[utf8]{inputenc}
\usepackage[small]{caption}
\usepackage{graphicx}
\usepackage{amsmath}
\usepackage{amsthm}
\usepackage{algorithm}
\usepackage{algorithmic}
\usepackage{epsfig}
\usepackage{amssymb}
\usepackage{pifont}
\usepackage{color,soul}

\usepackage[labelformat=simple]{subcaption}

\usepackage{multirow}

\usepackage{comment}
\usepackage{makecell}
\usepackage{mathrsfs}
\usepackage{booktabs}

\hypersetup{
    colorlinks=true,
    linkcolor=blue,
    filecolor=magenta,      
    urlcolor=magenta,
}

\newcommand{\etc}{\textit{etc}.}
\newcommand{\ie}{\textit{i.e.}, }
\newcommand{\eg}{\textit{e.g.}, }

\definecolor{mygreen}{RGB}{126, 171, 85}
\definecolor{myorange}{RGB}{222, 131, 68}

\newcommand{\brian}[1]{}
\renewcommand{\brian}[1]{{\color{red} BY: {#1}}}
\newcommand{\XD}[1]{}
\renewcommand{\XD}[1]{{\color{magenta} XD: {#1}}}
\newcommand{\EA}[1]{}
\renewcommand{\EA}[1]{{\color{blue} EA: {#1}}}

\title{Coupling Intent and Action for Pedestrian Crossing Behavior Prediction}

\author{
Yu Yao$^1$\and
Ella Atkins$^1$\and
Matthew Johnson Roberson$^1$\and
Ram Vasudevan$^1$\And
Xiaoxiao Du $^1$
\thanks{This work was supported by a grant from Ford Motor Company via the Ford-UM Alliance under award N028603. This material is based upon work supported by the Federal Highway Administration under contract number 693JJ319000009. Any options, findings, and conclusions or recommendations expressed in the this publication are those of the author(s) and do not necessarily reflect the views of the Federal Highway Administration.}
\affiliations
$^1$University of Michigan\\
\emails
\{brianyao, ematkins, mattjr, ramv, xiaodu\}@umich.edu
}

\begin{document}

\maketitle

\begin{abstract}
Accurate prediction of pedestrian crossing behaviors by autonomous vehicles can significantly improve traffic safety. Existing approaches often model pedestrian behaviors using trajectories or poses but do not offer a deeper semantic interpretation of a person’s actions or how actions influence a pedestrian's intention to cross in the future. In this work, we follow the neuroscience and psychological literature to define pedestrian crossing behavior as a combination of an unobserved inner will (a probabilistic representation of binary intent of crossing vs. not crossing) and a set of multi-class actions (\eg walking, standing, \etc). Intent generates actions, and the future actions in turn reflect the intent. We present a novel multi-task network that predicts future pedestrian actions and uses predicted future action as a prior to detect the present intent and action of the pedestrian. We also designed an attention relation network to incorporate external environmental contexts thus further improve intent and action detection performance. We evaluated our approach on two naturalistic driving datasets, PIE and JAAD, and extensive experiments show significantly improved and more explainable results for both intent detection and  action prediction over state-of-the-art approaches. \textit{Our code is available at: \url{https://github.com/umautobots/pedestrian_intent_action_detection}}
\end{abstract}

\section{Introduction}

Pedestrians are among the most exposed and vulnerable road users in traffic accidents and fatalities in urban areas \cite{cantillo2015modelling,yao2019unsupervised,yao2020dota}. In particular, pedestrians have been found more likely to be involved in crashes by vehicles and at greater risks during street crossing \cite{kim2019traffic}. Therefore, for the safe navigation and deployment of intelligent driving systems in populated environments, it is essential to accurately understand and predict pedestrian crossing behaviors. 

\begin{figure}[h!]
    \centering
    \includegraphics[width=1.0\columnwidth]{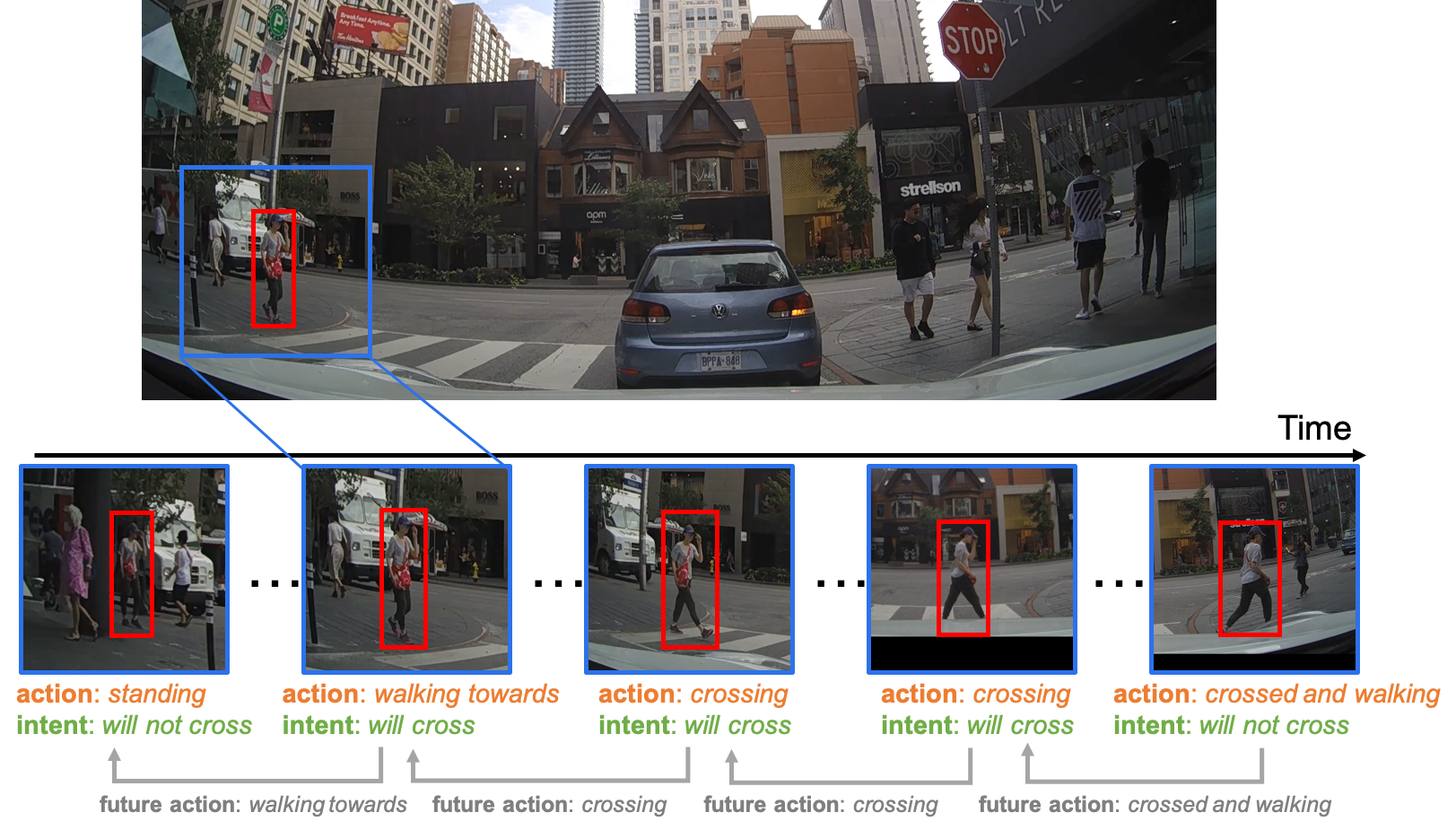}
    \caption{A sample video sequence illustrating our future action guided intent detection. A person goes through the sequence of first standing far from the intersection with no intent of crossing, and then starts going towards the crosswalk with crossing intent. The person then keeps crossing until reaching the state of crossed and walking. Successfully predicting future action of going towards and crossing can help detect the ``will cross'' intent.}
    \label{fig:intent_teaser}
    \vspace{-8pt}
\end{figure}

Many existing work that study pedestrian intention \cite{rehder2015goal,bai2015intention,karasev2016intent} referred to the term ``intention'' as goals in pedestrian trajectories and evaluated using trajectory or path location errors. However, this definition only models intent as motion but did not offer a deeper semantic interpretation of both the pedestrian's physical and mental activities. In this work, we follow neuroscience and psychology literature \cite{haggard2002voluntary,haggard2008human,schroder2014intention} and define ``intent'' as the inner consciousness or ``will'' of a person's decision to act. We hypothesize that a human's implicit intent affects the explicit present and future actions, and future actions in turn reflect the intent. Thus, we propose to leverage future action prediction as priors to perform online intent and action detection and prediction. We developed a multi-task network to encode pedestrian visual and location features, in addition to features from the future action predictor, to output probabilities of binary crossing intent (will cross or not) and multi-class semantic actions (\eg ``waiting'' or ``going towards the crosswalk''). We also developed an attentive relation module to extract object-level features (different from previously commonly used semantic maps) to incorporate environmental context in predicting pedestrian crossing behaviors. The main contributions of our work are two-fold:

\begin{itemize}
    \item We redefine the pedestrian behavior as a combination of a binary-class crossing intent and a multi-class semantic action and designed a novel multi-task model to detection and predict both jointly. We show that our method achieves superior (improved by $\sim5-25\%$, based on multiple metrics) and real-time ($<6ms$) intent and action prediction performance over the state-of-the-art on two in-the-wild benchmark driving datasets.
    \item Inspired by human visual learning, we designed an attentive relation network (ARN) to extract environmental information based on pre-detected traffic objects and their state. We show that this object-level relation feature is more effective for pedestrian intent detection task, compared to existing methods using semantic maps. Our ARN also offers interpretable attention results on the relations between the traffic objects in the urban scene and a pedestrian's crossing intent and actions.
\end{itemize}

\section{Related Work}
\paragraph{Intent Detection.} Intention detection algorithms offer an intuitive understanding to human behaviors and assist many application areas, including user purchase and consumption \cite{wang2020intention2basket,ding2018domain}, Electroencephalography (EEG)-based motion monitoring \cite{fang2020learning}, and human-robot interaction \cite{li2017implicit,chen2019antprophet}, among others. To model pedestrian behaviors in driving applications, prior work typically model intent as motion and use poses and/or trajectories to predict future goals or locations \cite{yao2019egocentric,fang2018pedestrian,yao2020bitrap,bouhsain2020pedestrian}. The convolutional neural network (CNN) have been used to detect pedestrian intents in videos \cite{saleh2019real}. The state-of-the-art for pedestrian crossing intent detection and most relevant to our work include  \textit{\textit{PIE$_{int}$}}~\cite{rasouli2019pie}, stacked fusion GRU (SF-GRU)~\cite{rasouli2020pedestrian} and multi-modal hybrid pedestrian action prediction (MMH-PAP)~\cite{rasouli2020multi}. \textit{PIE$_{int}$} uses a Convolutional LSTM network to encode past visual features and then concatenates encoded features and bounding boxes to predict pedestrian intent at the final frame. SF-GRU used pre-computed pose feature and two CNNs to extract pedestrian appearance and context feature separately. MMH-PAP used pre-computed semantic maps to extract environmental feature for intent prediction. However, none of these work explored the relation between action and intent, nor can they predict actions in addition to intent.

\paragraph{Action Recognition.} Recurrent neural networks (RNNs) and 3-D CNNs have been used for action recognition in TV shows, indoor videos and sports videos~\etc ~\cite{simonyan2014two,tran2015learning,wang2016temporal,carreira2017quo,feichtenhofer2019slowfast}. These methods typically make one prediction per video and are mainly used for off-line applications, such as video tagging and query. For real-time applications, online action detectors have been developed to classify video frames with only past observations~\cite{gao2017red,shou2018online,gao2019startnet}. However, none of these work explored pedestrian crossing actions, nor do they explore the causes of the action (the intent). The temporal recurrent network (TRN) \cite{xu2019temporal} is most relevant to our work as it provided evidence that predicting future action can better recognize the action that is currently occurring. Different from TRN which focused on video action, our work focuses on pedestrian behaviors and predicts future actions to assist intent detection. While TRN leveraged future action to detect current action, our method views future action as the consequence of current intent and uses predicted action as a prior for intent detection.

\section{Approach}

\begin{figure*}[htbp]
    \centering
    \includegraphics[width=0.9\textwidth]{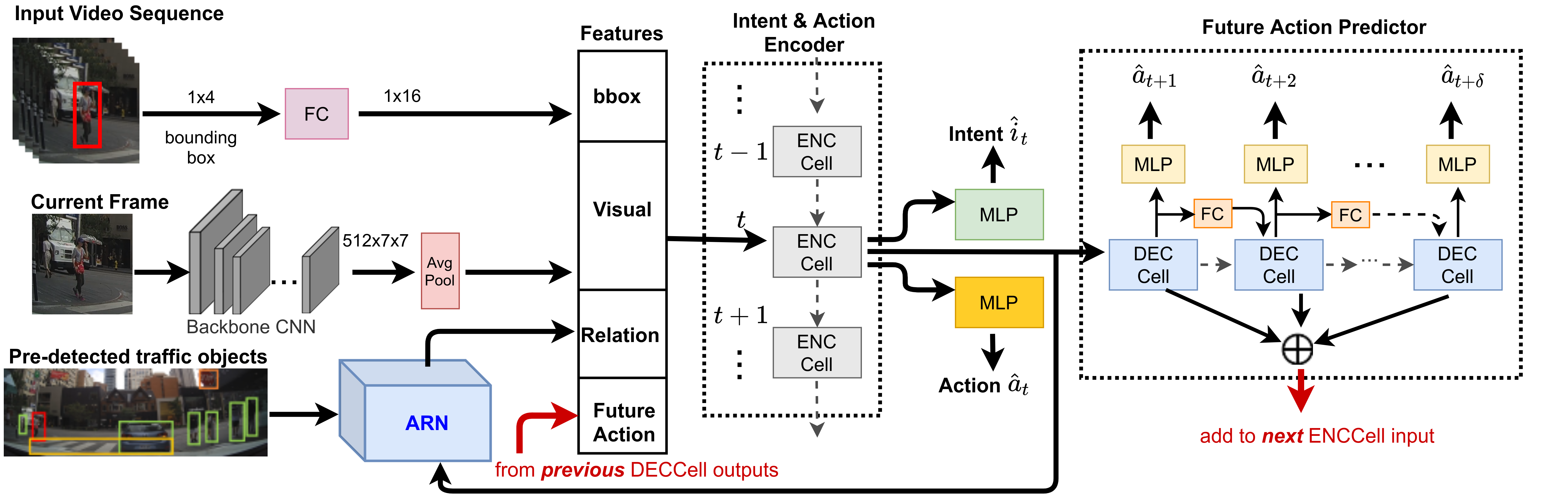}
    \caption{Our multi-task intent-action detection/prediction network.}
    \label{fig:intent_network}
    \vspace{-12pt}
\end{figure*}

Given a naturalistic driving dataset containing 2-D RGB traffic video clips,  let $\{O_1, O_2,...,O_t\}$ denote $t$ time steps of past observations (\ie image frames) in a sequence. The goal of our work is to detect the probability of a pedestrian's crossing intent $i_t \in [ 0,1]$ and the pedestrian's action $a_t \in \mathcal A$ at time $t$, and also predict future semantic actions $[a_{t+1},...,a_{t+\delta}]\in \mathcal A$, where $\mathcal A$ denotes the collection of possible actions related to crossing, as described in Section~\ref{sec:experiments}. The proposed architecture is illustrated in Figure~\ref{fig:intent_network}. The main part of our network contains a multi-task, encoder-decoder-based intent detection and action predictor, which outputs the detection results of the pedestrian crossing intent at the current time step and probabilistic forecast of future pedestrian behaviors (Section~\ref{ref:method_part1}). Additionally, an Attentive Relation Network (ARN) was developed to explore the surrounding environment for traffic objects that are in relation to the pedestrians (Section~\ref{ref:method_part2}). 

\subsection{Intent detection with future action prediction}
\label{ref:method_part1}
The multi-task intent and action prediction model consists of two parts: an intent-action encoder and a future action predictor. A multi-class cross-entropy-based loss function is used to optimize the network.
\paragraph{Intent-action encoder.} Given 2-D RGB traffic video clips, visual features were first extracted from image patches around each target pedestrian using a pre-trained deep convolutional neural network (CNN) following~\cite{rasouli2019pie}, as illustrated in the upper-left of Fig.~\ref{fig:intent_network}. An image patch is a square region defined by an enlarged pedestrian bounding box (``bbox'') that includes a pedestrian as well as some context information such as ground, curb, crosswalk, etc. Meanwhile, the bounding box coordinates of the image patch were passed through a fully-connected (FC) network to obtain location-scale embedding features. Then, the intent-action encoder (ENC) cells took the concatenated feature vectors as input and recurrently encoded the features over time. To ensure the neural network learns features that represent both crossing intent and more complicated semantic actions, a multi-layer perceptron (MLP)-based intent classifier and an action classifier were designed to use the encoder hidden state to predict the present intent and action simultaneously.

\paragraph{Action predictor.} Prior work such as the temporal recurrent neural network (TRN)~\cite{xu2019temporal} have shown that using future action predictions can improve the performance of classifying an action in the present. Inspired by this, our action predictor module was designed to utilize features from future action predictions and pass this information to the next ENCcell. In the action predictor module, the hidden state at each iteration was classified by an MLP network to predict the action at each future time step. Then, the average of all decoder hidden vectors was computed and used as the ``future" action feature. Such future action feature was concatenated with the visual and box features as described in the previous subsection to form the input to the next ENCcell. 

\paragraph{Multi-task loss.} The loss function of the proposed model consists of the binary cross entropy loss for the crossing/non-crossing intent detection, $H_B(\cdot)$, and the cross entropy loss for the multi-class action detection and prediction, $H(\cdot)$. The total loss can be written as
\begin{align}\label{eq:int_act_loss}
    \begin{split}
    \mathcal{L} = & \frac{1}{T}\sum_{t=1}^{T}\Big(\omega_1 H_B(\hat{i}_{t}, i_{t}) + \omega_2 H(\hat{a}_{t}, a_{t}) \\
    & + \omega_3 H([\hat{a}]_{t+1}^{t+\delta}, [a]_{t+1}^{t+\delta})\Big),    
    \end{split}
\end{align}
where $T$ is the total training sample length. Variables with and without $\hat{\cdot}$ indicate the predicted and ground truth values, respectively. The three loss terms refer to the intent detection loss for the current time step $t$, the action detection loss for the current time step $t$, and future action prediction loss for future time steps $t+1$ to $t+\delta$, respectively. The $\omega_1$, $\omega_2$ and $\omega_3$ are weighting parameters. In this work, the $\omega_1$ was designed to start with value $0$ and increase towards $1$ over the training process based on a sigmoid function of training iterations. This procedure ensures the model learns a reasonable action detection and prediction module, which can be used as a beneficial prior for the intent detector. For simplicity, $\omega_2$ and $\omega_3$ were both set to $1$ in training.

In the training stage, the network is updated in an online, supervised fashion, \ie given observation $O_1$, the network detects the current intention $i_1$, current action $a_1$, and predicts future actions $[a_2, a_3, ..., a_6]$ (if $\delta=5$, for example). Then, given the next observation $O_2$ and the encoded features from the previously predicted future actions $[a_2, a_3, ..., a_6]$, we detect the intention $i_2$ and again predicts future actions $[a_3, a_4, ..., a_7]$. This way, the predicted future actions of the pedestrian were leveraged and recurrently looped back as a prior to detect the present intent and action of that pedestrian, following our cognitive assumption that intent affects actions and actions in turn reflect the initial intent. This process goes on for all given training frames and the loss function was updated given the ground truth labels of intent and actions of training frames. In the testing stage, only image sequences were needed as input and the trained network generates the intent and action labels of all frames in the testing data.

\subsection{Attentive Relation Network (ARN)}
\label{ref:method_part2}
Dynamic environments have shown to impact pedestrian motion \cite{karasev2016intent}. To investigate and explore the relational features of the surrounding environments on the pedestrian behaviors, we designed an attentive relation network (ARN) that extracts informative features from traffic objects in the scene to add to the intent and action encoders to improve performance.  In this work, we consider six types of traffic objects that occurs most commonly in the scene, i.e., traffic agent neighbors (\eg pedestrians, cars, \etc); traffic lights; traffic signs; crosswalks; stations (\eg bus or train stops); and ego car. The feature vector of each object type was designed to reflect its relative position and state with respect to the target pedestrian. To be specific:

\paragraph{i) Traffic neighbor feature} $f_{ne} =[x^j_1-x^i_1, y^j_1-y^i_1, x^j_2-x^i_2, y^j_2-y^i_2]$, which represents the bounding box differences between the observed traffic participants (\eg vehicle, pedestrian, and cyclist) $j$ and the target pedestrian $i$. 

\paragraph{ii) Traffic light feature} $f_{tl} = [x^j_c-x^i_c, y^j_c-y^i_c, w^j, h^j, n_l^j, r^j]$, where $x^j_c-x^i_c$ and $ y^j_c-y^i_c$ are the coordinate differences between centers of traffic light $j$ and target pedestrian $i$, $w^j, h^j$ are the traffic light bounding box width and height, $n_l^j$ is the light type (\eg general or pedestrian light), and $r^j$ is the light status (\eg green, yellow or red).

\paragraph{iii) Traffic sign feature} $f_{ts} = [x^j_c-x^i_c, y^j_c-y^i_c, w^j, h^j, n_s^j]$, where the first four elements are similar to the traffic light feature, $n_s^j$ is the traffic sign type (\eg stop, crosswalk, \etc).

\paragraph{iv) Crosswalk feature} $f_{cw} = [x^j_1-x^i_{bc}, x^j_2-x^i_{bc}, y^j_1-y^i_{bc}, x^j_1, y^j_1, x^j_2, y^j_2]$, where $x^j_1-x^i_{bc}, x^j_2-x^i_{bc}, y^j_1-y^i_{bc}$ represent the coordinate differences between the bottom center of target pedestrian $i$ and the two bottom vertices of crosswalk $j$. We assume all pedestrians and crosswalks are on the same ground plane, thus, the bottom points better represent their depth in 3-D frame. $x^j_1, y^j_1, x^j_2, y^j_2$ are the crosswalk bounding box coordinates to show the crosswalk location information.

\paragraph{v) Station feature} $f_{st} = [x^j_1-x^i_{bc}, x^j_2-x^i_{bc}, y^j_1-y^i_{bc}, x^j_1, y^j_1, x^j_2, y^j_2]$, where the elements are similar to the crosswalk feature vector. 

\paragraph{vi) Ego motion feature:} $f_{eg}=[v, a, v_{yaw}, a_{yaw}]$, where $v$ and $a$ are the speed and acceleration, and $v_{yaw}, a_{yaw}$ are the yaw rate and yaw acceleration, collected from on-board diagnostics (OBD) of the data collection vehicle (ego vehicle).

\begin{figure}[h!]
    \centering
    \includegraphics[width=1.0\columnwidth, trim={3cm 0 0 0},clip]{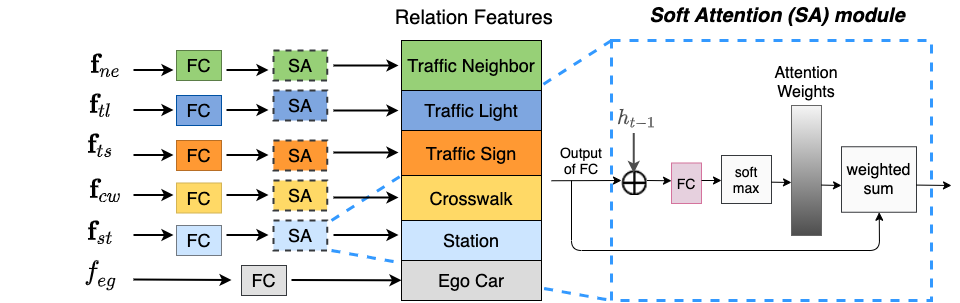}
    \caption{Our attentive relation network (ARN) with soft attention (SA). FC is fully-connected layer. Different colors correspond to different traffic objects. $\mathbf{f}_{(\cdot)}$ represents feature vectors for all objects, \eg $\mathbf{f}_{ne}=\left [ f^1_{ne}, ..., f^{N_{ne}}_{ne}\right ]$, where $N_{ne}$ is the total number of observed traffic participants. $h_{t-1}$ is the previous hidden state.}
    \label{fig:relation_net}
\end{figure}

The traffic object-specific features were extracted and computed as described above and an attentive relation network was designed to extract the relational features of all objects for a target pedestrian, as illustrated in Fig.~\ref{fig:relation_net}.  Each feature vector was first embedded to the same size by a corresponding FC network (\eg neighbor feature vector is embedded by $FC_{ne}$). For object types that may have multiple instances (\eg there can be more than one cars in the scene), a soft attention network (SAN) ~\cite{chen2017sca} was adopted to fuse the features. The SAN predicts a weighing factor for each instance, which represents the importance of that instance to the desired task, and then use a weighted sum to merge all instances. The ego car feature is not processed by a SAN, since current work assumes there is only one ego car at any given time. Then, the extract features of each traffic object type were concatenated to form the final relation feature $f_{rel} = [f_{ne}, f_{tl}, f_{ts}, f_{cw}, f_{st}, f_{eg}]$. Note that all elements of a feature vector were set to zeros if the corresponding object type was not observed in a scene. Finally, the relation features extracted through the ARN were concatenated with the visual, bounding box, and future action features as described in Section~\ref{ref:method_part1} to be passed together into the ENC network for intent and action encoding.

Compared to using visual features (\eg semantic maps) to represent environmental information as used in prior work \cite{rasouli2020pedestrian,rasouli2020multi}, our ARN directly uses pre-detected traffic object features, which is far more straightforward and informative. This is also inspired by human visual learning, as human drivers make decision based on observing objects rather than image pixels. The ARN is a very simple and fast network, since no deep CNNs were needed for visual feature extraction. 

\section{Experiments}
\label{sec:experiments}
We evaluate the effectiveness of our method for intent detection and action prediction on two publicly available naturalistic traffic video benchmarks, Pedestrian Intent Estimation (PIE) ~\cite{rasouli2019pie} and Joint Attention in Autonomous Driving (JAAD)~\cite{kotseruba2016jaad}. The PIE dataset was collected with an on-board camera covering six hours of driving footage. There are 1,842 pedestrians (880/243/719 for training/validation/test) with 2-D bounding boxes annotated at 30Hz with behavioral tags. Ego-vehicle velocity readings were provided using gyroscope readings from the camera. The JAAD dataset contains 346 videos with 686 pedestrians (188/32/126) captured from dashboard cameras, annotated at 30Hz with crossing intent annotations. Note that the original pedestrian action annotation in JAAD and PIE contains only binary categories, ``standing" and ``walking". To make the annotations more informative for intent detection, we leveraged the intent and crossing annotation to automatically augment the existing two into seven categories: standing, waiting, going towards (the crossing point), crossing, crossed and standing, crossed and walking, and other walking. 

\subsection{Experimental Settings}
\paragraph{Evaluation Metric.} Pedestrian crossing intent detection is traditionally defined as a binary classification problem (cross/not cross) and we include previous benchmark evaluation metrics including accuracy, F1 score, precision, and the receiver operating characteristic (ROC) area under curve (AUC) \cite{rasouli2020multi}. We introduce another metric in this work -- the difference between the average scores $s$ of all positive samples $P$ and all negative samples $N$, $\Delta_s = \frac{1}{|P|}\sum_{i\in P}s_i - \frac{1}{|N|}\sum_{i\in N}s_i$. This $\Delta_s$ metric reflects the margin between the detected positive and negative labels. The greater the $\Delta_s$, the larger the margin, thus, the more easily (better) the model can separate the crossing (positive) and not crossing (negative) classes. Note that accuracy, F1 score and precision can be biased towards the majority class if the test datasets were imbalanced (which is the case with both datasets used). 
Besides, these three metrics were computed at a hard threshold at 0.5 (score$>$0.5 is classified as positive, otherwise negative). On the other hand, both AUC and $\Delta_s$ metrics are less sensitive to the data imbalance and can more accurately reflect the different performance comparisons between methods. 

\paragraph{Implementation Details.} The proposed neural network model was implemented using PyTorch. We used gated recurrent units (GRU) with hidden size 128 for both ENCcell and DECcell, and a single-layer MLP as the classifier for intent detection, action detection and action prediction. The input image patch was prepared following~\cite{rasouli2019pie} to realize fair comparison, resulting in $224\times224$ image size.
An ImageNet-pre-trained VGG16 network acted as the backbone CNN to extract features from image patches~\cite{rasouli2019pie}. During training, the pedestrian trajectories were segmented to training samples with constant segment length $T$ in order to make batch training feasible. We applied a weighted sampler during training to realize balanced mini-batch. All models were trained with sample length $T=30$, prediction horizon $\delta=5$, learning rate $1e-5$, and batch size $128$ on a NVIDIA Tesla V100 GPU. 
During inference, the trained model was run on one test sequence at a time and the per-frame detection results were collected. 

\paragraph{Ablations.} We compared multiple variations of our method: The \textit{``I-only''} model is a basic model without action detection and prediction modules, which is similar to \textit{PIE$_{int}$}~\cite{rasouli2019pie} except that we used an average pooling feature extraction and a simpler GRU encoder father than the ConvLSTM network in \textit{PIE$_{int}$}; The \textit{``I+A''} model is a multi-task model with both intent and present action detection modules and the \textit{``I+A+F''} adds future action prediction; The \textit{``I+A+F+R''} or \textit{``full"} is our full model with the attentive relation network. We also present an \textit{``A-only''} and an \textit{``A+F''} ablations to show that future action prediction improves present action detection.

\subsection{Performance Evaluation and Analysis}
Two data sampling settings exist in previous work and we report results on both for direct and fair comparison: 1) \cite{rasouli2019pie} used all original data in PIE dataset, and 2) \cite{rasouli2020pedestrian,rasouli2020multi} sampled sequences from each track between 1-2 seconds up to the time of the crossing events. The second setting offers more future action changes than the first, since the first setting contains sequences long before the crossing event and actions tend not to change much until closer to the event of crossing.  

\begin{table}[htbp]
    \centering
    \caption{Intent and action detection results on the PIE dataset using the original data setting. \textbf{Bold} is best and \underline{underline} is second best.} 
    \resizebox{1.0\columnwidth}{!} {
    \begin{tabular}{l|ccccc|c}
        \toprule
        \multirow{2}{*}{Method} &\multicolumn{5}{c|}{intent} & action\\
        \cmidrule{2-7}
        & Acc$\uparrow$ & F1$\uparrow$ & Prec & AUC$\uparrow$  $\uparrow$ & $\Delta_s \uparrow$ & mAP$\uparrow$\\ 
        \midrule
        Naive - 1 & \textbf{0.82} & \textbf{0.90} & 0.82 & 0.50 & 0.00 & N/A\\
        \textit{PIE$_{int}$}~\cite{rasouli2019pie} & \underline{0.79} & 0.87 & 0.86 & 0.73 & 0.06 & N/A\\
        \midrule
        Ours (\textit{I-only}) & \textbf{0.82} & \textbf{0.90} & 0.86 & 0.72 & 0.07 & N/A\\ 
        Ours (\textit{A-only}) & N/A & N/A & N/A & N/A & N/A & 0.18 \\
        Ours (\textit{I+A}) & 0.74 & 0.83 & 0.90 & 0.76 & 0.11 & 0.21\\
        Ours (\textit{A+F}) & N/A & N/A & N/A & N/A & N/A & 0.20\\
        Ours (\textit{I+A+F}) & \underline{0.79} & 0.87 & \underline{0.90} & \underline{0.78} & \underline{0.16} & \underline{0.23} \\
        Ours (\textit{I+A+F+R (full)}) & \textbf{0.82} & \underline{0.88} & \textbf{0.94} & \textbf{0.83} & \textbf{0.41} & \textbf{0.24}\\
        \bottomrule
    \end{tabular}}
    \label{tab:intent_pie}
\end{table}

\begin{table}[h!]
    \centering
    \caption{Results on PIE and JAAD datasets using ``event-to-crossing'' setting. Detection precision (Prec) is presented and $\Delta_s$ was removed to compare with baselines using their metrics.} 
    \resizebox{1.0\columnwidth}{!} {
    \begin{tabular}{l|cccc|cccc}
        \toprule
        \multirow{2}{*}{Method} & \multicolumn{4}{c|}{PIE} & \multicolumn{4}{c}{JAAD} \\
        \cmidrule{2-9}
        & Acc & F1  & AUC & Prec & Acc & F1  & AUC & Prec \\
        \midrule
        Naive-1 or 0 & 0.81 & \textbf{0.90} & 0.50 & 0.81 & 0.79 & \textbf{0.88} & 0.50 & \textbf{0.79} \\
        I3D~\cite{carreira2017quo} &0.63  & 0.42 & 0.58 & 0.37 & 0.79 & 0.49 & 0.71 & 0.42 \\
        SF-GRU~\cite{rasouli2020pedestrian} & 0.87 & 0.78 & 0.85 & 0.74 & 0.83 & 0.59 & 0.79 & 0.50 \\
        MMH-PAP~\cite{rasouli2020multi} & \textbf{0.89} & 0.81 & \textbf{0.88} & 0.77 & 0.84 & 0.62 & 0.80 & 0.54 \\
        \midrule
        Ours (\textit{full}) & 0.84 & \textbf{0.90} & \textbf{0.88} & \textbf{0.96} & \textbf{0.87} & 0.70 & \textbf{0.92} & 0.66 \\
        \bottomrule
    \end{tabular}}
    \label{tab:pie_jaad_results}
\end{table}

\subsubsection{Quantitative Results on the Original Data Setting}
The top three rows of Table~\ref{tab:intent_pie} shows the intent detection results of the baseline methods. \textit{Naive-1} achieves 0.82 accuracy, 0.90 F1 score and 0.82 precision, even higher than the previously state-of-the-art \textit{PIE$_{int}$} method due to the imbalanced test set of PIE dataset. This observation verifies our previous discussion on the inadequacy of using only accuracy, F1 score and precision as evaluation metrics.  The $\Delta_s$ of naive methods was $0.00$ because it cannot distinguish crossing and not crossing classes, while  $\Delta_s$ score of \textit{PIE$_{int}$} is slightly higher at 0.06.  

The bottom rows of Table~\ref{tab:intent_pie} below the middle line show ablation results on the variations of our proposed method. The \textit{I-only} baseline, with action detection and prediction removed, achieves similar results with \textit{PIE$_{int}$} in terms of AUC and $\Delta_s$, indicating our choice of an average pooling and simple GRU network with less parameters can perform similarly to the ConvLSTM encoder in \textit{PIE$_{int}$}. The \textit{I+A} model significantly increased AUC and $\Delta_s$, indicating an improved capability in distinguishing between crossing and not crossing intent. However, its lower accuracy and F1 scores show that it missed some crossing intents. One explanation is that more negative (not crossing) samples previously misclassified by the \textit{I-only} model are now correctly classified when the action type is known in the \textit{I+A}  model. For example, a person standing close to the curb but not facing the road can be regarded as just ``standing" there instead of ``waiting". The \textit{I+A} model recognizes such differences and will correctly classify this example as ``not crossing", while the \textit{I-only} model may lose that information. The \textit{I+A+F} model further improves the AUC and $\Delta_s$ as well the accuracy and F1 scores from the \textit{I+A} ablation. This verifies our assumption that future action is a more straightforward indicator to intent, \eg a high confidence on ``waiting" or ``crossing" action indicates higher probability that a pedestrian has crossing intent. The higher $\Delta_s$ results show that adding future action stream helps the model to better distinguish different crossing intents. Our \textit{full} model, \textit{I+A+F+R}, significantly improves the accuracy, AUC and $\Delta_s$, indicating the effectiveness of adding the environmental information from ARN in detecting intent. 

\subsubsection{Quantitative Results on the ``Event of Crossing'' Setting}

Table~\ref{tab:pie_jaad_results} shows the results of our method on PIE and JAAD following training and evaluation method in~\cite{rasouli2020pedestrian,rasouli2020multi} considering only sequences between 1-2s to the event of crossing. We again observed that the \textit{Naive-1} (for PIE) or \textit{-0} (for JAAD) methods achieved higher accuracy, F1 score and precision on both datasets due to the imbalanced testing samples, since PIE and JAAD are dominated by positive and negative samples respectively. AUC is a more reliable metric in this case. Our method outperforms the state-of-the-art SF-GRU and MMH-PAP methods and achieved the highest 0.88 AUC on PIE and 0.92 AUC on JAAD datasets, which indicated that predicting the change of action through our future-action-conditioned module indeed helps improve the detection of the crossing intent. 

\paragraph{Ablation Study on ARN:} Table~\ref{tab:per_type} studies the impact of each different traffic object type in the ARN. As shown, adding traffic lights and/or crosswalks significantly improves the results, while the rest of the objects by themselves do not seem to help much. This result indicated that the ARN network learned that the two major factors of determining pedestrian crossing intent include traffic lights and crosswalks, which fits human intuition. 

\begin{table}[htbp]
    \centering
    \caption{Ablation study on ARN on the PIE dataset. The first row shows our I+A+F model, while the following are models with each single traffic object type, including ego motion. The bottom row shows our full model with all traffic object types in PIE dataset.}
    \resizebox{1.0\columnwidth}{!} {
    \begin{tabular}{c|c|c|c|c|c|cccc}
        \toprule
        \makecell{Traffic\\ neighbor} & \makecell{Cross\\walk} & \makecell{Traffic\\ light} & \makecell{Traffic\\ sign} & Station & \makecell{Ego\\ car} & Acc & F1 & AUC & $\Delta_s$ \\
        \midrule
         -& -& -&- & -& -& 0.79 & \underline{0.87} & 0.78 & 0.16 \\
        \midrule 
        \ding{52}&&&&&& 0.73 & 0.83 & 0.70 & 0.16 \\
        \midrule
        &\ding{52}&&&&& 0.77 & 0.84 & \textbf{ 0.83} & 0.31 \\
        \midrule 
        &&\ding{52}&&&& \underline{0.81} & \textbf{0.88} & \underline{0.81} & \underline{0.39} \\
        \midrule 
        &&&\ding{52}&&& 0.78 & 0.86 & 0.73 & 0.12\\
        \midrule 
        &&&&\ding{52}&& 0.71 & 0.80 & 0.77 & 0.12\\
        \midrule 
        &&&&&\ding{52} & 0.79 & \underline{0.87} & 0.73 & 0.10\\
        \midrule 
        \ding{52}&\ding{52}&\ding{52}&\ding{52}&\ding{52}&\ding{52} & \textbf{0.82} & \textbf{0.88} & \textbf{0.83} & \textbf{0.41}\\
        \bottomrule
    \end{tabular}}
    \label{tab:per_type}
\end{table}

\begin{figure}[h!]
    \centering
    \begin{subfigure}[h]{0.49\textwidth}
        \includegraphics[width=1.0\columnwidth]{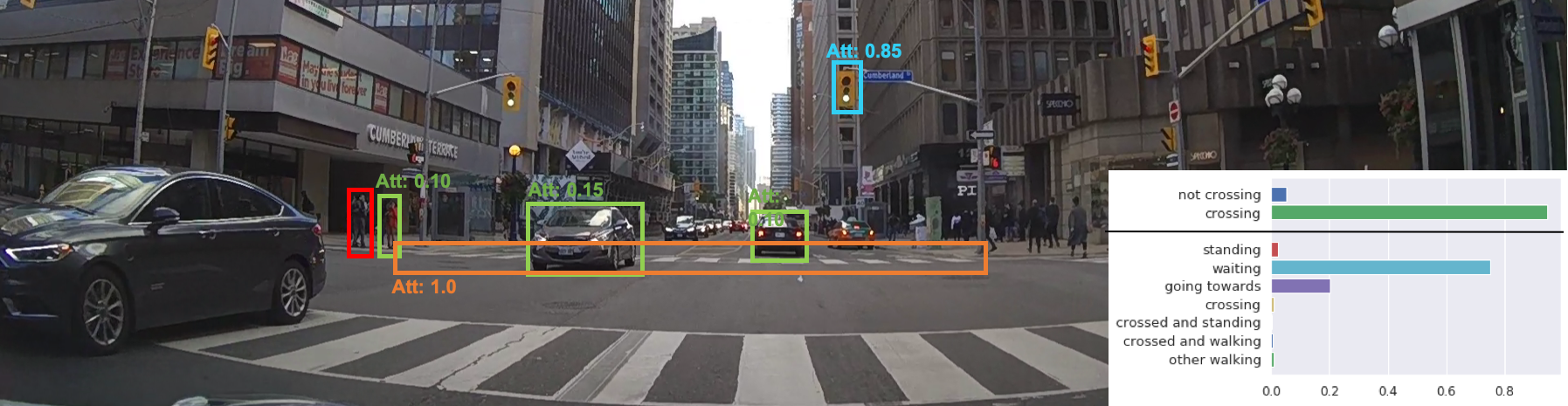}
        \caption{A pedestrian waiting to cross at an intersection. The predicted crossing scores are: 0.65 (\textit{I-only}), 0.70 (\textit{I+A}), 0.75 (\textit{I+A+F}) and \textbf{0.95 (\textit{full})}.}
        \label{fig:result_1_with_att}
    \end{subfigure}
    
    \begin{subfigure}[h]{0.49\textwidth}
        \includegraphics[width=1.0\columnwidth]{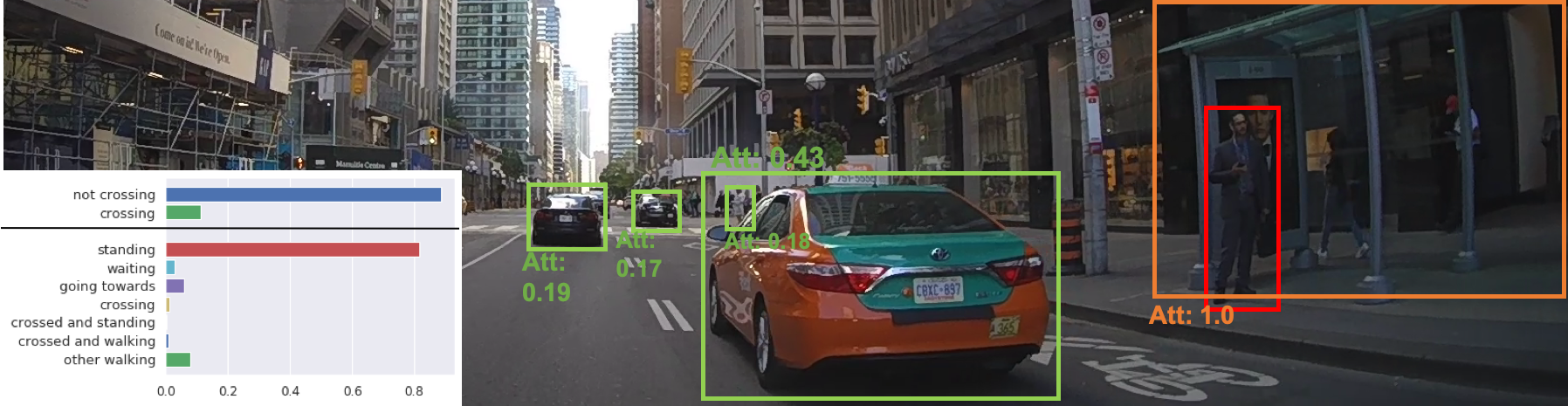}
        \caption{A pedestrian waiting for bus at a bus stop without crossing intent. The predicted non-crossing scores are: 0.31 (\textit{I-only}), 0.36 (\textit{I+A}), 0.35 (\textit{I+A+F}) and \textbf{0.89 (\textit{full})}.}
        \label{fig:result_2_with_att}
    \end{subfigure}
    
    \begin{subfigure}[h]{0.49\textwidth}
        \includegraphics[width=1.0\columnwidth]{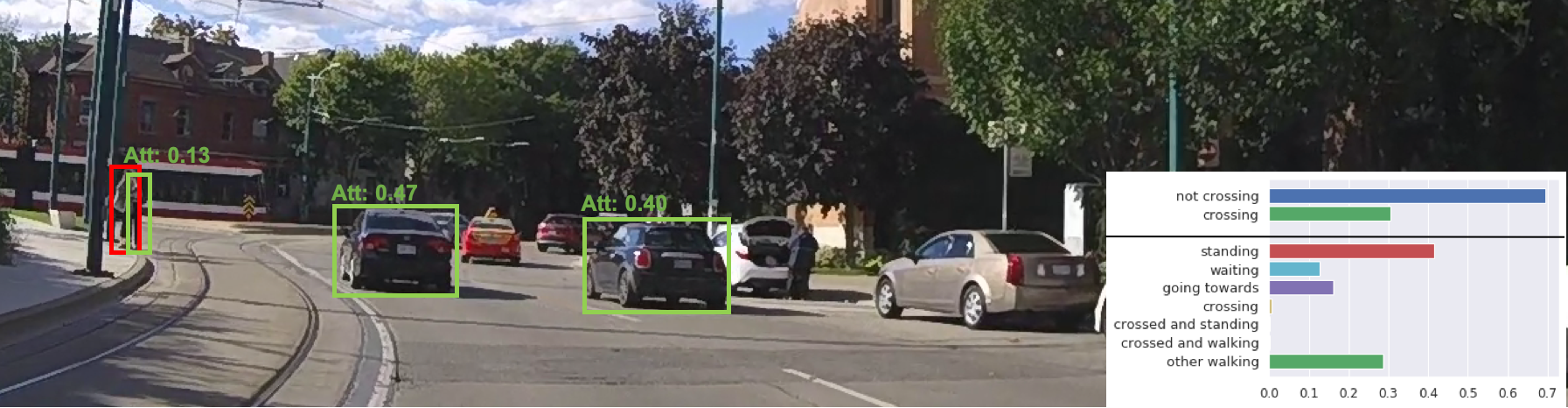}
        \caption{A pedestrian waiting to cross a street without traffic light/sign or crosswalk. The predicted crossing scores are: 0.65 (\textit{I-only}), 0.68 (\textit{I+A}), \textbf{0.80 (\textit{I+A+F})} and 0.31 (\textit{full}).}
        \label{fig:result_3_with_att}
    \end{subfigure}
    \caption{Qualitative examples of our full model on PIE dataset. The target pedestrians are shown by red boxes. Up to top-5 attentive traffic participants, traffic lights and infrastructures from ARN are shown in green, cyan and orange boxes together with the intermediate attention probabilities in our model. The predicted crossing-intent scores and action scores are presented in bar charts.}
    \label{fig:qualitative}
\end{figure}

\subsubsection{Qualitative Examples} 
Fig.~\ref{fig:qualitative} shows three examples of running our \textit{full} model on the PIE dataset. As shown in Fig~\ref{fig:result_1_with_att}, our \textit{I+A+F} model has higher confidence in ``crossing'' prediction than \textit{I+A} and \textit{I-only}, demonstrating again the effectiveness of predicting future actions as the prior of current intent detection. Our \textit{full} model predicts highest score for ``crossing" intent and the ``waiting" action by paying higher attention to the front crosswalk, traffic light, and the traffic cars and the nearby pedestrian that has same intent and action. The pedestrian traffic light was not detected, thus, no attention was assigned. Fig.~\ref{fig:result_2_with_att} shows an example where a pedestrian is waiting for bus at a bus stop. All the models without ARN predicted low score for non-crossing intent, while our \textit{full} model improves the result significantly since it captures the bus stop (orange box) and also the taxi (green box) in front of it.

Fig~\ref{fig:result_3_with_att} shows a ``failure" case of our \textit{full} method. In this scene, two pedestrians are waiting to cross the street without crosswalk, traffic light or traffic sign. Our \textit{I+A+F} model predicted the highest score for crossing intent, since it focused on the context near the target pedestrian and the temporal actions in the future. The \textit{full} model, on the contrary, predicted low scores, since it did not find the infrastructures related to a crossing event (\eg crosswalk, traffic light, traffic sign \etc). Moreover, the \textit{full} model paid higher attention to the two cars than the nearby pedestrian who was also waiting to cross, making it more difficult to recognize the crossing intent of the target pedestrian. This challenging case indicated that while the current model with ARN is successful at incorporating relation features, it sometimes lacks the ability to recognize when to rely on relation feature and when to switch rely on target pedestrian's feature, which is an interesting future work area to further explore. Our current ARN also shares the same drawback with prior supervised methods in that it cannot extract features from environmental objects that were unseen during training. Semi-supervised or unsupervised methods may be developed to solve this issue.

\subsection{Discussion on Computation Time}
\label{sec:computation_time}
We report the computation time to show the efficiency of our models. All experiments were conducted on a single NVIDIA Tesla V100 GPU. The mean and standard deviation ($\mu\pm\sigma$) of the per-frame inference time of our models are $0.5\pm0.1ms$ (\textit{I-only}), $0.5\pm0.1ms$ (\textit{I+A}), $2.2\pm0.1ms$ (\textit{I+A+F}), $6.0\pm1.2ms$ (\textit{full}), respectively. Our \textit{full} model, being slightly slower than its ablations, is still fast enough for real-time applications. The larger standard deviation was due to the fact that the number of traffic objects varied largely in different scenes. As a comparison, we ran the original implementation of previous $PIE_{int}$ on the same machine and observed a higher $9\pm0.5ms$ inference time.

\section{Conclusion}
In this work, we proposed a multi-task action and intent prediction network for pedestrian crossing behavior given naturalistic driving videos. We showed through extensive experiments that our method outperforms the state-of-the-art by a large margin and can jointly and recurrently predict present and future actions and intent for every pedestrian in every frame, in real-time. Through our novel future action predictor module, we verified our hypothesis that there is a relation between intent and action, that intent affects actions and that future action is a useful prior to accurately predict intent. We also studied the relations among various traffic objects and showed that our ARN module assigns attention to nearby objects that affect crossing intent and improve the performance.

\clearpage
\small
\bibliographystyle{named}
\bibliography{ijcai21}

\end{document}